\documentclass[10pt]{article}

\usepackage{ranlp}
\usepackage{epsfig}
\usepackage{amsfonts}

\def\TB#1{{\sf TreeBoost[$#1$]}}
\def\TreeBoost{{\sf TreeBoost}}
\def\STUMPS{{\sf Stumps}}
\def\aWL{{\sf WeakLearner}}    % El nom de la funcio que apren un Weak Learner
\def\X{\mathcal X}

\def\ig{\!=\!}              % La igualtat ajustada
\def\pert{\!\in\!}          % La pertinenca ajustada
\def\lbd{[\![}     
\def\rbd{]\!]}

\title{Boosting Trees for Anti-Spam Email Filtering}

\author{Xavier Carreras \and Llu\'\i s M\`arquez\\
   TALP Research Center\\
   LSI Department\\
   Universitat Polit\`ecnica de Catalunya (UPC)\\
   Jordi Girona Salgado 1--3\\
   Barcelona E-08034, Catalonia\\
   \texttt{\{carreras,lluism\}@lsi.upc.es}}

\begin{document}

\maketitle

%----------------------------------------- A b s t r a c t
\begin{abstract} 
  This paper describes a set of comparative experiments for the
  problem of automatically filtering unwanted electronic mail
  messages. Several variants of the AdaBoost algorithm with
  confidence--rated predictions \cite{schapire99} have been applied,
  which differ in the complexity of the base learners considered.  Two
  main conclusions can be drawn from our experiments: a) The
  boosting--based methods clearly outperform the baseline learning
  algorithms (Naive Bayes and Induction of Decision Trees) on the PU1
  corpus, achieving very high levels of the $F_1$ measure; b)
  Increasing the complexity of the base learners allows to obtain
  better ``high--precision'' classifiers, which is a very important
  issue when misclassification costs are considered.
\end{abstract}

%\subsection*{Keywords}
%spam, information filtering, information retrieval, 
%text categorization, machine learning for natural language processing

%----------------------------------------- I n t r o d u c t i o n
\section{Introduction}
Spam-mail filtering is the problem of automatically filtering unwanted
electronic mail messages. The term ``spam mail'' is also commonly
referred to as ``junk mail'' or ``unsolicited commercial mail''.
Nowadays, the problem has achieved a big impact since bulk emailers
take advantage of the great popularity of the electronic mail
communication channel for indiscriminately flooding email accounts
with unwanted advertisements. The major factors that contribute to the
proliferation of unsolicited spam email are the following two: 1) bulk
email is inexpensive to send, and 2) pseudonyms are inexpensive to
obtain
\cite{cranor98}. On the contrary, individuals may waste a large amount
of time transferring unwanted messages to their computers and sorting
through those messages once transferred, to the point that they may be
likely to become overwhelmed by spam.
 
Automatic IR methods are well suited for addressing this problem ,
since spam messages can be distinguished from the ``legitimate'' email
messages because of their particular form, vocabulary and word
patterns, which can be found in the header or body of the messages.

The spam filtering problem can be seen as a particular instance 
of the Text Categorization problem (TC), in which only two 
classes are possible: {\it spam} and {\it legitimate}. However, since
one is the opposite of the other, it also can be seen as the problem of
identifying a single class, {\it spam}. In this way, the evaluation of
automatic spam filtering systems can be done by using common measures 
of IR (precision, recall, etc.).
Another important issue is the relative importance between the two types of
possible misclassifications: While an automated filter that misses a
small percentage of spam may be acceptable to most users, fewer
people are likely to accept a filter that incorrectly
identifies a small percentage of legitimate mail as spam, especially
if this implies the automatic discarding of the misclassifed
legitimate messages. This problem suggests the consideration of
misclassification costs for the learning and evaluation of spam filter
systems. 

In recent years, a vast amount of techniques have been applied to TC,
achieving impressive performances in some cases.  Some of the
top--performing methods are Ensembles of Decision Trees
\cite{weiss99}, Support Vector Machines \cite{joachims98}, Boosting
\cite{schapire00} and Instance--based Learning \cite{yang99b}.  Spam
filtering has also been treated as a particular case of TC. In 
\cite{cohen96} a method based on TF-IDF weighting and the rule
learning algorithm RIPPER is used to classify and filter
email. \cite{sahami98} used the Naive Bayes approach to filter spam
email. \cite{drucker99} compared Support Vector Machines (SVM),
boosting of C4.5 trees, RIPPER and Rocchio, concluding that SVM's and
boosting are the top--performing methods and suggesting that SVM's are
slightly better in distinguishing the two types of
misclassification. In \cite{androus00a} Sahami's Naive Bayes is
compared against the TiMBL Memory-based learner. In \cite{androus00b}
the same authors present a new public data set which might become a
standard benchmark corpus, and introduce cost-sensitive evaluation
measures.

In this paper, we show that the AdaBoost algorithm with
confidence--rated predictions is a very well suited algorithm for
addressing the spam filtering problem. We have obtained very accurate
classifiers on the PU1 corpus, and we have observed that the algorithm
is very robust to overfitting. Another advantage of using AdaBoost is
that no prior feature filtering is needed since it is able to
efficiently manage large feature sets (of tens of thousands).

In the second part of the paper we show how increasing the
expressiveness of the base learners can be exploited for obtaining the
``high--precision'' filters that are needed for real user
applications.  We have evaluated the results of such filters using the
measures introduced in \cite{androus00b}, which take into account the
misclassification costs. We have substantially improved the results
mentioned in that work.

%%% We think that this is the more important contribution of this paper.

The paper is organized as follows: Section \ref{s-MLmethods} is
devoted to explain the AdaBoost learning algorithm and the variants
used in the comparative experiments. In section \ref{s-setting} the
setting is presented in detail, including the corpus and the
experimental methodology used. Section \ref{s-experiments} reports the
experiments carried out and the results obtained. Finally, section
\ref{s-conclusions} concludes and outlines some lines for further 
research.

%----------------------------------------- L e a r n i n g   A l g o r i t h m s
\section{The AdaBoost Algorithm}
\label{s-MLmethods}
In this section the AdaBoost algorithm \cite{schapire99} is
described, restricting to the case of binary classification.

The purpose of boosting is to find a highly accurate classification
rule by combining many {\it weak rules} (or weak hypotheses), each of
which may be only moderately accurate. It is assumed the existence of
a separate procedure called the \aWL\ for acquiring the weak
hypotheses. The boosting algorithm finds a set of weak hypotheses by
calling the weak learner repeatedly in a series of $T$ rounds. These
weak hypotheses are then linearly combined into a single rule called
the {\it combined hypothesis}.

Let $S=\{(x_1,y_1),\dots,(x_m,y_m)\}$ be the set of $m$ training
examples, where each instance $x_i$ belongs to a instance space $\X$
and $y_i\pert\{-1,+1\}$ is the class or label associated to $x_i$.
The goal of the learning is to produce a function of the form
$f:\X\rightarrow\mathbb{R}$, such that, for any example $x$, the sign
of $f(x)$ is interpreted as the predicted class ($-1$ or $+1$), and
the magnitude $|f(x)|$ is interpreted as a measure of confidence in
the prediction. Such a function can be used either for classifying or
ranking new unseen examples.

The pseudo--code of AdaBoost is presented in Figure~\ref{f-abmh}.  It
maintains a vector of weights as a distribution $D$ over examples. The
goal of the \aWL\ algorithm is to find a weak hypothesis with
moderately low error with respect to these weights.  Initially, the
distribution $D_1$ is uniform, but the boosting algorithm
exponentially updates the weights on each round to force the weak
learner to concentrate on the examples which are hardest to predict by
the preceding hypotheses.

\begin{figure}[htb]
\hspace*{0.2cm}\rule[0pt]{7.7cm}{0.4pt}\vspace*{2mm}\\
{\small
\hspace*{2mm}  {\bf procedure} AdaBoost ({\bf in}: $S=\{(x_i,y_i)\}_{i=1}^m$)\smallskip\\
\hspace*{2mm}      {\tt \#\#\#} $S$ is the set of training examples\\
\hspace*{2mm}       {\tt \#\#\#} Initialize distribution $D_1$ (for all $i$, $1\leq i\leq m$)\smallskip\\
\hspace*{5mm}       $D_1(i)=1/m$\smallskip\\
\hspace*{2mm}       {\bf for } $t$:=1 {\bf to} $T$ {\bf do}\smallskip\\
\hspace*{5mm}         {\tt \#\#\#} Get the weak hypothesis $h_t: \X\rightarrow\mathbb{R}$\smallskip\\
\hspace*{5mm}         $h_t$ = \aWL\,($X,D_t$);\smallskip\\
\hspace*{5mm}         Choose $\alpha_t\in\mathbb{R}$;\smallskip\\
\hspace*{5mm}         {\tt \#\#\#} Update distribution $D_t$ (for all $i$, $1\leq i\leq m$)\smallskip\\
\hspace*{8mm}           $D_{t+1}(i)\ig {\displaystyle\frac{D_t(i){\rm exp}(-\alpha_t y_ih_t(x_i))}{Z_t}}$\smallskip\\
\hspace*{5mm}           {\tt \#\#\#} $Z_t$ is chosen so that $D_{t+1}$ will be a distribution\smallskip\\
\hspace*{2mm}       {\bf end-for}\\
\hspace*{2mm}       {\bf return} the combined hypothesis: $f(x)={\displaystyle\sum_{t=1}^{T}\alpha_t h_t(x)}$\\
\hspace*{2mm}       {\bf end} AdaBoost\\
}
\hspace*{0.2cm}\rule[0pt]{7.7cm}{0.4pt}\vspace*{0mm}
\caption{The AdaBoost algorithm}
\label{f-abmh}
\end{figure}  

More precisely, let $D_t$ be the distribution at round $t$, and
$h_t\,:\,\X\rightarrow\mathbb{R}$ the weak rule acquired according to
$D_t$. In this setting, weak hypotheses $h_t(x)$ also make
real--valued confidence--rated predictions (i.e., the sign of $h_t(x)$
is the predicted class, and $|h_t(x)|$ is interpreted as a measure of
confidence in the prediction).  A parameter $\alpha_t$ is then chosen
and the distribution $D_t$ is updated. The choice of $\alpha_t$ will
be determined by the type of weak learner (see next section).  In the
typical case that $\alpha_t$ is positive, the updating function
decreases (or increases) the weights $D_t(i)$ for which $h_t$ makes a
good (or bad) prediction, and this variation is proportional to the
confidence $|h_t(x_i)|$.  The final hypothesis, $f$, computes its
predictions using a weighted vote of the weak hypotheses.

In \cite{schapire99} it is proven that the training error of
the AdaBoost algorithm on the training set (i.e. the fraction of
training examples $i$ for which the sign of $f(x_i)$ differs from
$y_i$) is at most $\prod_{t=1}^T Z_t$, where $Z_t$ is the
normalization factor computed on round $t$. This upper bound is used
in guiding the design of both the parameter $\alpha_t$ and the \aWL\
algorithm, which attempts to find a weak hypothesis $h_t$ that
minimizes $Z_t$.

%------------------------------
\subsection{Learning weak rules}

In \cite{schapire99} three different variants of AdaBoost.MH are
defined, corresponding to three different methods for choosing the
$\alpha_t$ values and calculating the predictions of the weak
hypotheses. In this work we concentrate on AdaBoost\ {\it with
real--valued predictions} since it is the one that has achieved the
best results in the Text Categorization domain \cite{schapire00}.

According to this setting, weak hypotheses are simple rules with
real--valued predictions. Such simple rules test the value of a
boolean predicate and make a prediction based on that value. The
predicates used refer to the presence of a certain word in the text,
e.g. ``the word {\it money} appears in the message''.  Formally, based
on a given predicate $p$, our interest lies on weak hypotheses $h$
which make predictions of the form: $$
h(x)=\left\{\begin{array}{ll}c_{0}&\ {\rm if\ } p {\rm\ holds\ in\ }
x\\ c_{1}&\ {\rm otherwise} \end{array}\right.  $$
\noindent where the $c_0$ and $c_1$ are real numbers.

For a given predicate $p$, the values $c_0$ and $c_1$ are calculated
as follows.  Let $X_1$ be the subset of examples for which the
predicate $p$ holds and let $X_0$ be the subset of examples for which
the predicate $p$ does not hold. Let $\lbd\pi\rbd$, for any predicate
$\pi$, be 1 if $\pi$ holds and 0 otherwise.  Given the current
distribution $D_t$, the following real numbers are calculated for
$j\pert\{0,1\}$, and for $b\pert\{+1,-1\}$: $$
W_b^{j}={\displaystyle\sum_{i=1}^mD_t(i)\lbd x_i\in X_j\land
y_i=b\rbd}\,.  $$

That is, $W_{b}^{j}$ is the weight, with respect to the distribution
$D_t$, of the training examples in partition $X_j$ which are of class
$b$.  As it is shown in \cite{schapire99} $Z_t$ is minimized for a
particular predicate by choosing:
\begin{equation}
  c_{j}=\frac{1}{2}\,{\rm ln}\left(\frac{W_{+1}^{j}}{W_{-1}^{j}}\right)\,.
\end{equation}
\noindent and by setting $\alpha_t=1$. These settings imply that:
\begin{equation}
Z_t=2{\displaystyle\sum_{j\in\{0,1\}}\sqrt{W_{+1}^{j}W_{-1}^{j}}}\,.
\end{equation}
Thus, the predicate $p$ chosen is that for which the value of 
$Z_t$ is smallest. 

Very small or zero values for the parameters $W_b^{j}$ cause $c_{j}$
predictions to be large or infinite in magnitude.  In practice, such
large predictions may cause numerical problems to the algorithm, and
seem to increase the tendency to overfit. As suggested
in~\cite{schapire00}, the smoothed values for $c_{j}$ have been
considered.

It is important to see that the so far presented weak rules can be
directly seen as decision trees with a single internal node (which
tests the value of a boolean predicate) and two leaf nodes that give
the real-valued predictions for the two possible outcomes of the test.
These extremely simple decision trees are sometimes called {\it
decision stumps}.  In turn, the boolean predicates can be seen as
binary features (we will use the word {\it feature\/} instead of {\it
predicate}\/ from now on), thus, the already described criterion for
finding the best weak rule (or the best feature) can be seen as a
natural splitting criterion and used for performing decision--tree
induction \cite{schapire99}.

Following the idea suggested in \cite{schapire99} we have extended the
\aWL\ algorithm to induce arbitrarily deep decision trees. The
splitting criterion used consists in choosing the feature that
minimizes equation (2), while the predictions at the leaves of the
boosted trees are given by equation (1). Note that the general
AdaBoost procedure remains unchanged.

In this paper, we will denote as \TreeBoost\ the AdaBoost.MH algorithm
including the extended \aWL. \TB{d} will stand for a learned
classifier with weak rules of depth $d$. As a special case, \TB{0}
will be denoted as \STUMPS.

%----------------------------------------- S e t t i n g
\section{Setting}
\label{s-setting}

\subsection{Domain of Application}
We have evaluated our system on the PU1 benchmark corpus\footnote{PU1
Corpus is freely available from the publications section of {\tt
http://www.iit.demokritos.gr/$\sim$ionandr}} for the anti-spam email
filtering problem.  It consists of 1,099 messages: 481 of them are
spam and the remaining 618 are legitimate. The corpus is presented
partitioned into 10 folds of the same size which maintain the
distribution of spam messages \cite{androus00b}. All our experiments
have been performed using 10-fold cross-validation.

The feature set of the corpus is a bag-of-words model. Four versions
are available: with or without stemming, and with or without stop-word
removal. The experiments reported in this paper have been performed
using the non-stemming non-stop-word-removal version, which consists
in a set of 26,449 features.

%At the moment, four versions of the PU1 corpus are available,
%differing in the type of features used. In all cases, the feature set
%is a bag-of-words. Two resources were used for
%the generation: a stop-word list and a lemmatizer. The
%enabling/disabling of these two resources in the generation results in 
%the following four versions: 
%\begin{itemize}
%\item BARE: No resources used. The features are the encrypted
%lower-cased words. Consists of 26,449 features.
%\item LEMM: Lemmatizer used; no stop-word removal. Consists of 23,312 features.
%\item STOP: Stop-word list used; no lemmatizer. Consists of 26,275 features.
%\item LEMM STOP: Both lemmatizer and stop-word list used. Consists of
%23,137 features.
%\end{itemize}

\subsection{Experimental Methodology}
\paragraph*{Evaluation Measures.}
Measures for evaluating the spam filtering system are introduced here.
Let $S$ and $L$ be the number of spam and legitimate messages in the
corpus, respectively; let $S_+$ denote the number of spam messages
that are correctly classified by a system, and $S_-$ the number of
spam messages misclassified as legitimate. In the same way, let $L_+$
and $L_-$ be the number of legitimate messages classified by a system
as spam and legitimate, respectively. These four values form a
contingency table which summarizes the behaviour of a system. The
widely-used measures precision (p), recall (r) and $F_\beta$ are defined as
follows:
%\vspace*{-1mm}
\begin{displaymath}
p = \frac{ S_+ }{ S_+ + L_+ }
\quad
r = \frac{ S_+ }{ S_+ + S_- }
\quad
F_\beta = \frac{(\beta^2 + 1)pr}{\beta^2 p + r}
\end{displaymath}

%\begin{displaymath}
%Accuracy: acc = \frac{ L_- + S_+ }{ L  + S }
%\quad
%Error: err = \frac{ L_+ + S_- }{ L  + S }
%\end{displaymath}

The $F_\beta$ measure combines precision and recall, and with $\beta =
1$ gives equal weigth to the combined measures. Additionally, some
experiments in the paper will also consider the accuracy measure ($acc
= \frac{ L_- + S_+ }{ L + S }$).

A way to distinguish the two types of misclassification is the
use of utility measures \cite{lewis95} used 
in the TREC evaluations \cite{trec99}. In this general measure, 
positions in the contingency table are associated loss values,
$\lambda_{S+}$, $\lambda_{S-}$, $\lambda_{L+}$, $\lambda_{L-}$,  
which indicate how desirable are the outcomes, according to a
user-defined scenario. The overall performance of a system in terms
of the utility is $S_+ \lambda_{S+} + S_- \lambda_{S-} + L_+
\lambda_{L+} + L_- \lambda_{L-}$.

Androutsopoulos et al. \cite{androus00b} propose particular scenarios
in which misclassifying a legitimate message as spam is $\lambda$
times more costly than the symmetric misclassification. In terms of
utility, these scenarios can be translated to $\lambda_{S+}=0$,
$\lambda_{S-}=-1$, $\lambda_{L+} = -\lambda$ and $\lambda_{L-} =
0$. They also introduce the \emph{weighted accuracy} (WAcc) measure,
a version of accuracy sensitive to $\lambda$-cost:
%\vspace*{-1mm}
\begin{displaymath}
WAcc = \frac{ \lambda \cdot L_- + S_+ }{ \lambda \cdot L + S }
%\quad
%WErr = \frac{ \lambda \cdot L_+ + S_- }{ \lambda \cdot L + S }
\end{displaymath}

When evaluating filtering systems, this measure suffers from the same
problems as standard accuracy \cite{yang99a}. Despite this fact, we
will use it for comparison purposes.

\subsection{Baseline Algorithms}
In order to compare our boosting methods against other techniques, we
include the following two baseline measures:
\begin{itemize} 
%\item No Filter. This simple method always classifies a
%message as legitimate. It can be seen as the most frequent class
%criterion, provided that legitimate mails are more frequent 
%than spam mails.
\item Decision Trees. Standard TDIDT learning algorithm, using the RLM
distance-based function for the feature selection. See \cite{marquez99a} for complete
details about the particular implementation.
\item Naive Bayes. We include the best results on the PU1 Corpus
reported in \cite{androus00b}, corresponding to a Naive Bayes
classifier.
\end{itemize}

%----------------------------------------- E x p e r i m e n t s

\section{Experiments}
\label{s-experiments}
This section explains the set of experiments carried out. As said in
section \ref{s-setting}, all experiments work with the PU1 Corpus.

\subsection{Comparing methods on the corpus}

The purpose of our first experiment is to show the general performance of 
boosting methods in the spam-filtering domain. Six AdaBoost
classifiers have been learned, setting the depth of the weak rules from 0 to
5; we denote each classifier as \TB{d}, where $d$ stands for the depth of
the weak rules; as a particular case, we denote the \TB{0} classifier as
\STUMPS. Each version of \TreeBoost\ has been learned for up to 2,500 weak rules.

\begin{figure}[htb]
\centering
\epsfig{file=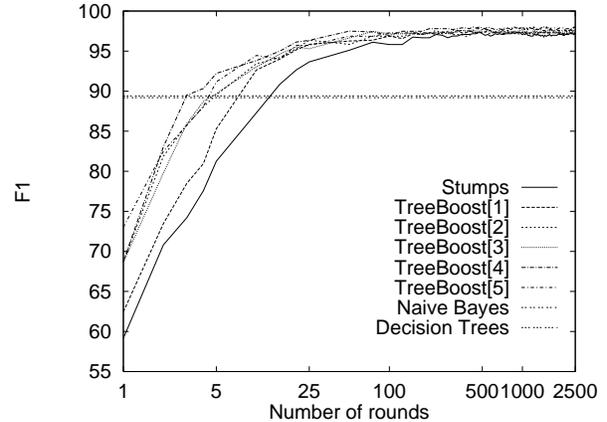, width=8cm}
\caption{$F_1$ measure of \STUMPS\ and \TB{d}, for increasing number of rounds}
\label{fig:generalperformance}
\end{figure}

Figure \ref{fig:generalperformance} shows the $F_1$ measure of
each classifier, as a function of the number of rounds used. In this
plot, there are also the obtained rates of the baseline algorithms. It
can be seen that \TreeBoost\ clearly outperforms
the baseline algorithms. The experiment also shows
that, above a certain number of rounds, all \TreeBoost\ versions
achieve consistent good results, and that there is no overfitting in
the process. After 150 rounds of boosting, all versions reach an $F_1$
value above 97\%. It can be noticed that the deeper the weak rules,
the smaller the number of rounds needed to achieve good
performance. This is not surprising, since deeper weak
rules handle much more information. Additionally, the figure shows
that different number of rounds produce slight variations in the error rate.

A concrete value for the $T$ parameter of the \TreeBoost\ learning
algorithm must be given, in order to obtain real classifiers and to be
able to make comparisons between the different versions of \TreeBoost\
and baseline methods.  To our knowledge, it is still unclear what is
the best way for choosing $T$. We have estimated the $T$ parameter in
a validation set built for the task, with the following procedure: a)
For each trial in the cross-validation experiment, 8 of the 9 training
subsets are used to learn up to 2,500 rounds of boosting, and the one
remaining is used as the validation set for testing the classifier
with respect to the number of rounds, in steps of 25. b) The outputs
of all classifiers are used to compute the $F_1$ measure. c) The
minimum $T$ for which the $F_1$ is maximum is chosen as the estimated
optimal number of rounds for all classifiers.

%\textbf{$T$ tuning procedure:} Given a training set $F$ composed of 10 folds:
%\begin{itemize}
%\item For each fold $F_i$, $i = 1 .. 10$:
%\begin{itemize}
%\item Randomly select one fold, $validation_{i}$, among the 9 training
%folds $F_j$, $j \neq i$.
%\item Learn a classifier $TB_i$, for 2,500 rounds,
%using the remaining 8 folds $F_j$, with $j \neq $i and $j \neq validation_{i}$.
%\end{itemize}
%\item For $T$ values from 0 to 2,500 (in steps of 25): 
%\begin{itemize}
%\item Test each classifier $TB_{i}$ on its set
%$validation_{i}$, using $T$ weak rules. 
%\item Compute the F$1$ measure, based on the predictions of all folds.
%\end{itemize} 
%\item Choose the minimum value of $T$ for which error rate is minimum.
%\end{itemize}

Table \ref{table:results} presents the results of all classifiers. For
each one, we include the number of rounds (estimated in the validation
set), recall, precision, $F_1$ and the maximum $F_1$
achieved over the 2,500 rounds learned.  According to the results,
boosting classifiers clearly outperform the other algorithms. Only
Naive Bayes achieves a precision (95.11\%) slightly lower than the
obtained by boosting classifiers (the worse is 97.48\%); however, its
recall at this point is much lower.

\begin{table}[htb]
\centering
\begin{tabular}{|l|c|c|c|c|c|} \hline
 & $T$ & Recall & Prec. & $F_1$ & $F^{max}_1$ \\ \hline \hline
%No Filter & - & 56.23 & 0 & $\infty$ & - \\ \hline
N. Bayes & - & 83.98 & 95.11 & 89.19 & - \\ \hline
D. Trees & - & 89.81 & 88.71 & 89.25 & - \\ \hline \hline
\STUMPS & 525 & 96.47 & 97.48 & 96.97 & 97.39 \\ \hline
\TB{1} & 525 & 96.88 & 97.90 & 97.39 & 97.60\\ \hline
\TB{2} & 725 & 96.67 & 98.31 & 97.48 & 97.59 \\ \hline
\TB{3} & 675 & 96.88 & 97.90 & 97.39 & 97.81 \\ \hline
\TB{4} & 450 & 97.09 & 98.73 & 97.90 & 98.01\\ \hline
\TB{5} & 550 & 96.88 & 98.52 & 97.69 & 98.12\\ \hline 
\end{tabular}
\caption{Performance of all classifiers}
\label{table:results}
\end{table}

Accuracy results have been compared using the 10-fold
cross-valida\-ted paired $t$ test. Boosting
classifiers perform significantly better than Decision
Trees.\footnote{Since we do not own the Naive Bayes classifiers, no
tests have been ran; but presumably boosting methods are also
significantly better.} On the contrary, no significant differences can
be observed between the different versions of
\TreeBoost. More interestingly, it can be 
noticed that accuracy and precision rates slightly increase with the
expressiveness of the weak rules, and that this improvement does not
affect the recall rate. This fact will be exploited in the following
experiments.

\subsection{High-Precision classifiers}

This section is devoted to evaluate \TreeBoost\ in high-precision
scenarios, where only a very low (or null) proportion of legitimate to 
spam misclassifications is allowed.

\begin{figure}[hbt]
\centering
\begin{tabular}{cc}
\STUMPS & \TB{1} \\
\hspace*{-.5cm}\epsfig{file=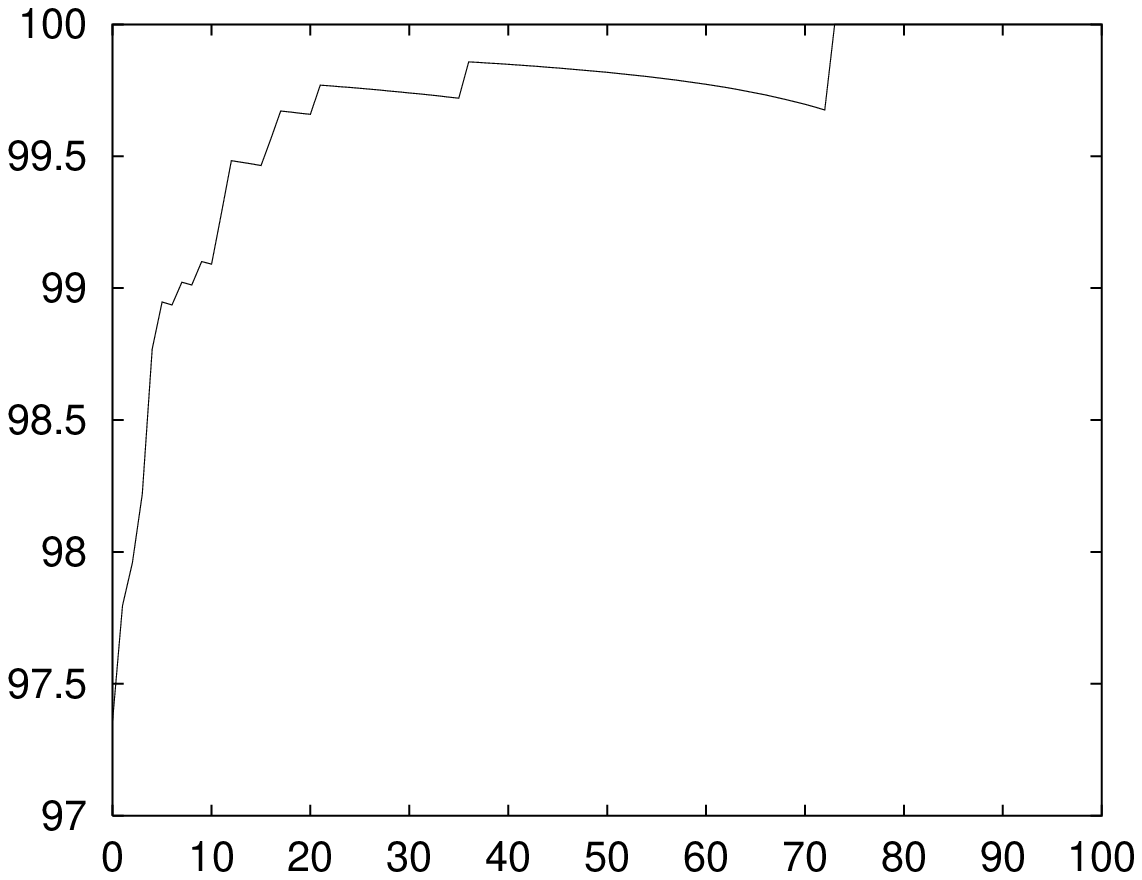, width=4.4cm} &
\hspace*{-.7cm}\epsfig{file=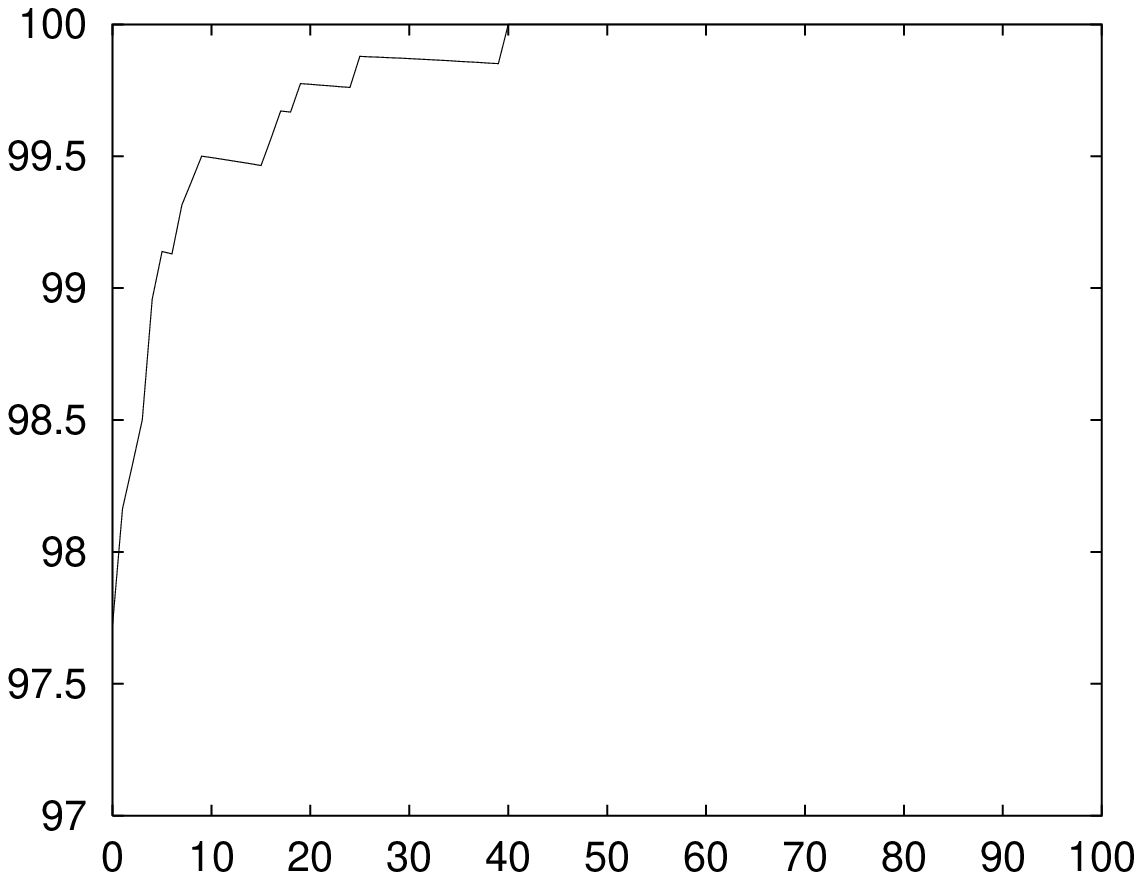, width=4.4cm} \\
\TB{2} & \TB{3} \\
\hspace*{-.5cm}\epsfig{file=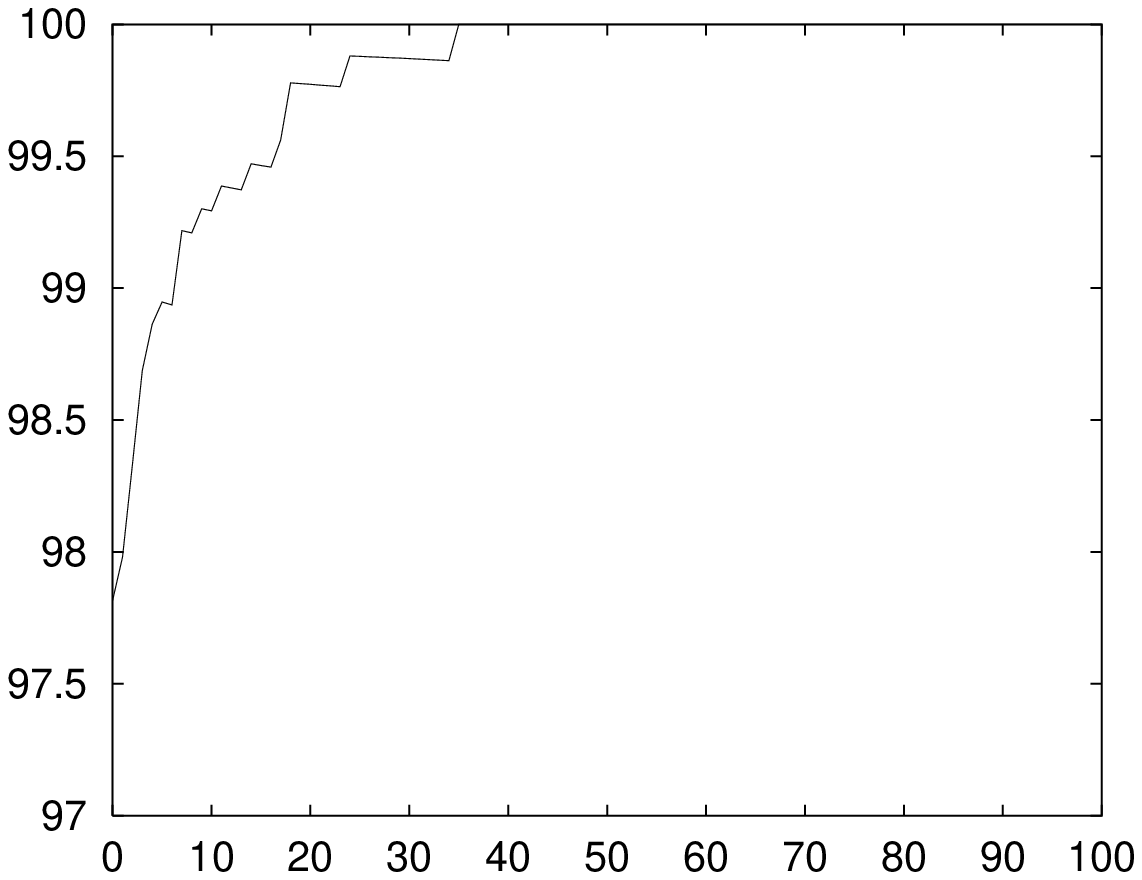, width=4.4cm} &
\hspace*{-.7cm}\epsfig{file=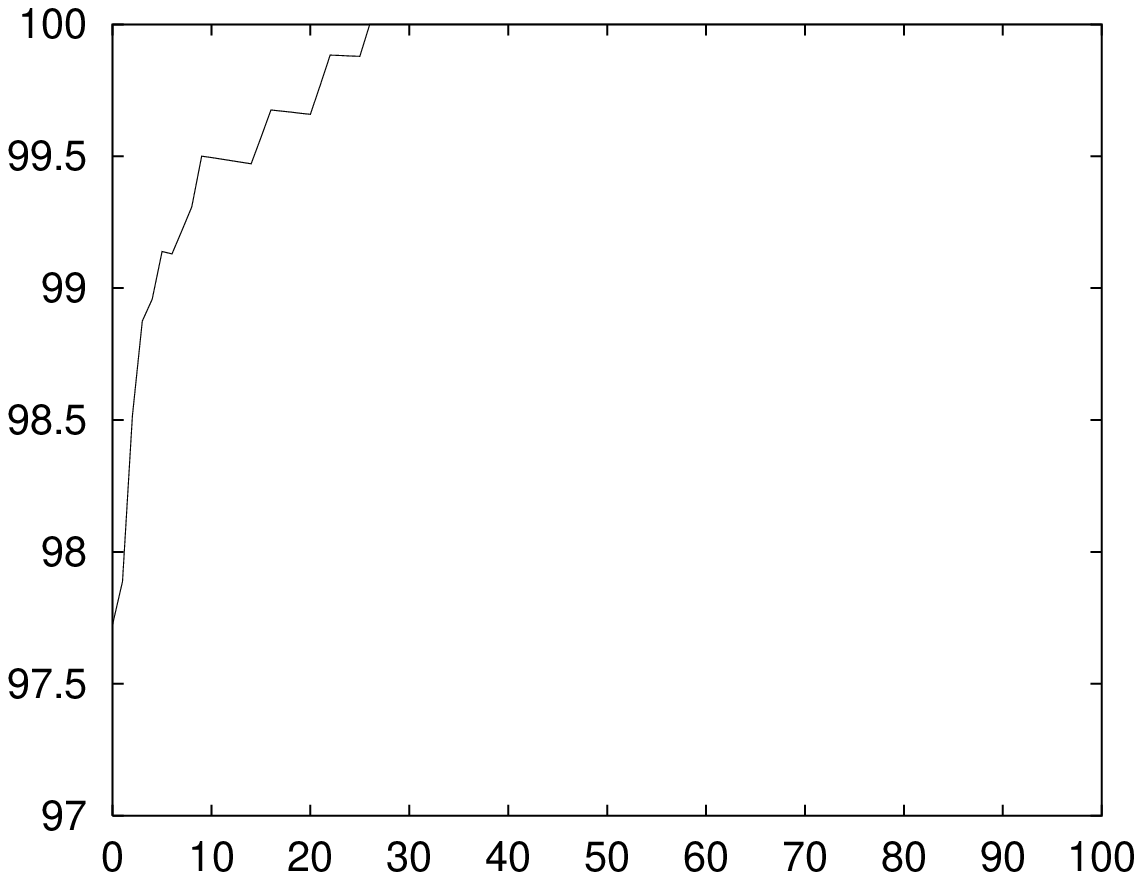, width=4.4cm} \\
\TB{4} & \TB{5} \\
\hspace*{-.5cm}\epsfig{file=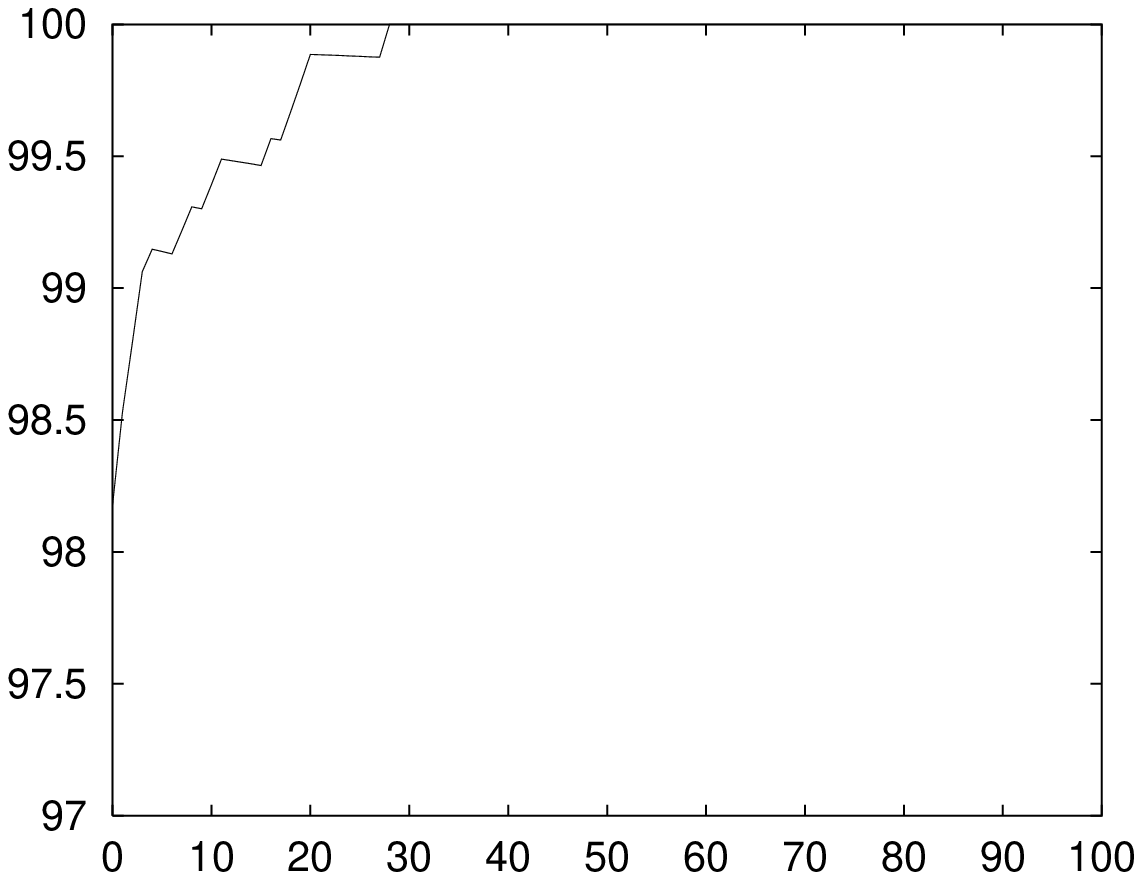, width=4.4cm} &
\hspace*{-.7cm}\epsfig{file=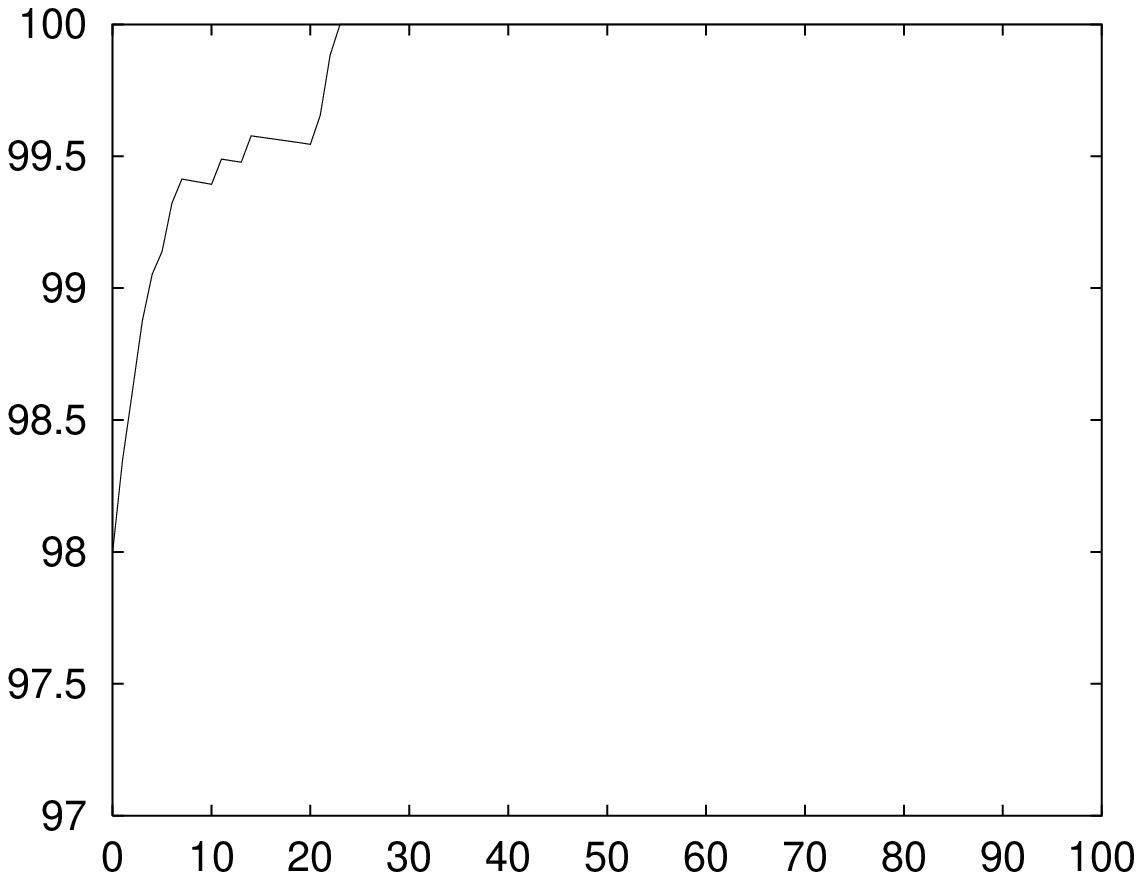, width=4.4cm} \\
\end{tabular}
\caption{Rejection curves for all \TreeBoost\ classifiers. $x$ axis:
percentage of rejected predictions; $y$ axis: accuracy.}
\label{fig:rejcurves}
\end{figure}

\paragraph*{Rejection Curves.}
We start by evaluating if the confidence of a prediction, i.e., the
magnitude of the prediction, is a good indicator of the quality of the
prediction.  For this purpose, rejection curves are computed for each
classifier. The procedure to compute a rejection curve is the
following: For several points $p$ between 0 and 100, reject the $p\%$
of the predictions whose confidences score lowest, both positive or
negative, and compute the accuracy of the remaining $(100-p)\%$
predictions. This results in higher accuracy values as long as $p$
increases. Figure \ref{fig:rejcurves} plots the rejection curves
computed for the six learned classifiers. The following conclusions
can be drawn:
\begin{itemize}
\item The confidence of a prediction is a good indicator of its quality.
\item Depth of weak rules greatly improves the quality of the predictions. Whereas
\STUMPS\ needs to reject the 73\% of the less confident examples to achieve a 100\% of
accuracy, \TB{5} only needs 23\%. In other words, deeper \TreeBoost\
filters concentrate the misclassified examples closer to the decision
threshold.
\item The previous fact has important consequences for a potential
final email filtering application, with the following specification:
Messages whose prediction confidence is greater than a threshold $\tau$
are automatically classified: spam messages are blocked and legitimate
messages are delivered to the user. Messages whose prediction confidence
is lower than $\tau$ are stored in a special fold for dubious
messages. The user has to verify if these are legitimate
messages. This specification is suitable for having automatic filters
with different degrees of strictness (i.e., different values for the
$\tau$ parameter). $\tau$ values could be tuned using a validation
set.
\end{itemize}

\paragraph*{Cost--Sensitive Evaluation.}

In this section, \TreeBoost\ classifiers are evaluated using the
$\lambda$-cost measures introduced in section \ref{s-setting}.  Three
scenarios of strictness are presented in \cite{androus00b}: a) No cost
considered, corresponding to $\lambda = 1$; b) Semi-automatic
scenario, for a moderately accurate filter, giving $\lambda = 9$; and
c) Completely automatic scenario, for a very high accurate filter,
assigning $\lambda = 999$. As noted in section
\ref{s-setting}, we will consider these scenarios as particular
utility matrices.

%Moreover, to include the $\lambda$ factor in their 
%Naive Bayes classification scheme, they define the following classification rule: 
%A message is labelled as a spam message if and only if its predicted 
%probability of being spam is $\lambda$ times greater than its predicted 
%probability of being legitimate.
%\begin{displaymath}
%\frac{ P( message is spam) }{P( message is legitimate )} > \lambda
%\end{displaymath}

In \cite{schapire98a} a modification of the Ada\-Boost algorithm for
handling general utility matrices is presented. The idea is to
initialize the weight distribution of examples according to the given
utility matrix, and then run the learning algorithm as usual. We have
performed experiments with this setting, but the results are not
convincing: only the initial rounds of boosting are affected by the
initialization based on utility; after a number of rounds, the
performance seems to be like if no utility had been considered. Since
our procedure for tuning the number of rounds can not determine when
the initial stage ends, we have rejected this approach. We think that
the modification of the AdaBoost algorithm should also consider the
weight update.

Another approach consists in adjusting the decision threshold
$\theta$. In a default scenario, corresponding to $\lambda=1$, an
example is classified as spam if its prediction is greater than 0; in
this case, $\theta = 0$.  Increasing the value of $\theta$ results in
a higher precision classifier. \cite{lewis95} presented a procedure
for calculating the optimal decision threshold for a system, given an
arbitrary utility matrix. The procedure is valid only when the system
outputs probabilities, so the prediction scores resulting from the
boosting classifications should be mapped to probabilities.  A method
for estimating probabilities given the output of AdaBoost is suggested
in \cite{friedman98}, using a logistic function. Initial experiments
with this function have not worked properly, because relatively low
predictions are sent to extreme probability values. A possible
solution would be to scale down the predictions before applying the
probability estimate; however, it can be observed that prediction
scores grow with both the number and the depth of the used weak
rules. Since many parameters are involved in this scaling, we have
rejected the probability estimation of predictions.

Alternatively, we make our classification scheme sensitive to
$\lambda$ factor by tuning the $\theta$ parameter to the value which
maximizes the weighted accuracy measure. Once more, the concrete value
for $\theta$ is obtained using a validation set, in which several
values for the parameter are tested. Table \ref{tab:costevaluation}
summarizes the results obtained from such procedures, giving $\lambda$
factor values of 9 and 999. Results obtained in \cite{androus00b} are
also reported.

Again, \TreeBoost\ clearly outperforms the baseline methods.  With
$\lambda = 9$, very high-precision rates are achieved, maintaining
considerably high recall rates; it seems that the depth of
\TreeBoost\ slightly improves the performance, although no significant 
differences can be achieved.  For $\lambda = 999$, precision rates of
100\% (which is the implicit goal in this scenario) are achieved,
except for \STUMPS, maintaining fair levels of recall. However, recall
rates are slightly unstable with respect to the depth of \TreeBoost\,
varying from 64.45\% to 76.30\%. Our impression is that high values in
the $\lambda$ factor seem to introduce instability in the evaluation,
which becomes oversensitive to outliers. In this particular corpus
(which contains 1,099 examples), \emph{weighted accuracy} does not
seem to work properly when giving $\lambda$ values of 999, since the
misclassification of only one legitimate message leads to score worse
than if any email had been filtered (this would give
$WAcc=99.92\%$). Moreover, for 100\% precision values, the recall
variation from 0\% to 100\% only affects the measure in 0.08 units.

\begin{table*}[htb]
\centering
\begin{tabular}{|l||c|c|c|c||c|c|c|c|} \hline
 & \multicolumn{4}{|c||}{$\lambda = 9$} & \multicolumn{4}{|c|}{$\lambda = 999$} \\ \hline
Method & $\theta$ & Recall & Prec. & WAcc & $\theta$ & Recall &
Prec. & WAcc \\ \hline \hline
%No Filter & - & 0 & $\infty$ & 92.04 & - & 0 & $\infty$ & 99.92  \\ \hline
Naive Bayes & - & 78.77 & 96.65 & 96.38 & - & 46.96 & 98.80 & 99.47 \\ 
\hline
%Decision Trees & ?? & ?? & ?? & ?? & ?? & ?? & ?? & ?? \\ \hline \hline
\STUMPS & 9.162 & 89.60 & 99.08 & 98.58 & 16.298 &  84.20 &  99.51 &
99.66 \\ \hline
\TB{1} & 10.2 & 93.55 & 98.71 & 98.59 & 46.87 & 74.43 & 100 & 99.98 \\ \hline
\TB{2} & 24.156 & 93.76 & 98.90 & 98.76 & 96.717 & 76.30 & 100 & 99.98 \\ \hline
\TB{3} & 45.165 & 91.48 & 99.32 & 98.87 & 126.388 & 74.01 & 100 & 99.98 \\ \hline
\TB{4} & 19.063 & 94.80 & 99.35 & 99.14 & 123.097 & 64.45 & 100 & 99.97 \\ \hline
\TB{5} & 37.379 & 93.97 &  99.12 & 98.92 & 177.909 & 66.53 & 100 & 99.97 \\ \hline
\end{tabular}
\caption{Cost--sensitive evaluation results}
\label{tab:costevaluation}
\end{table*}

\begin{figure}[hbt]
\centering
\begin{tabular}{cc}
\STUMPS & \TB{1} \\
\hspace*{-0.5cm}\epsfig{file=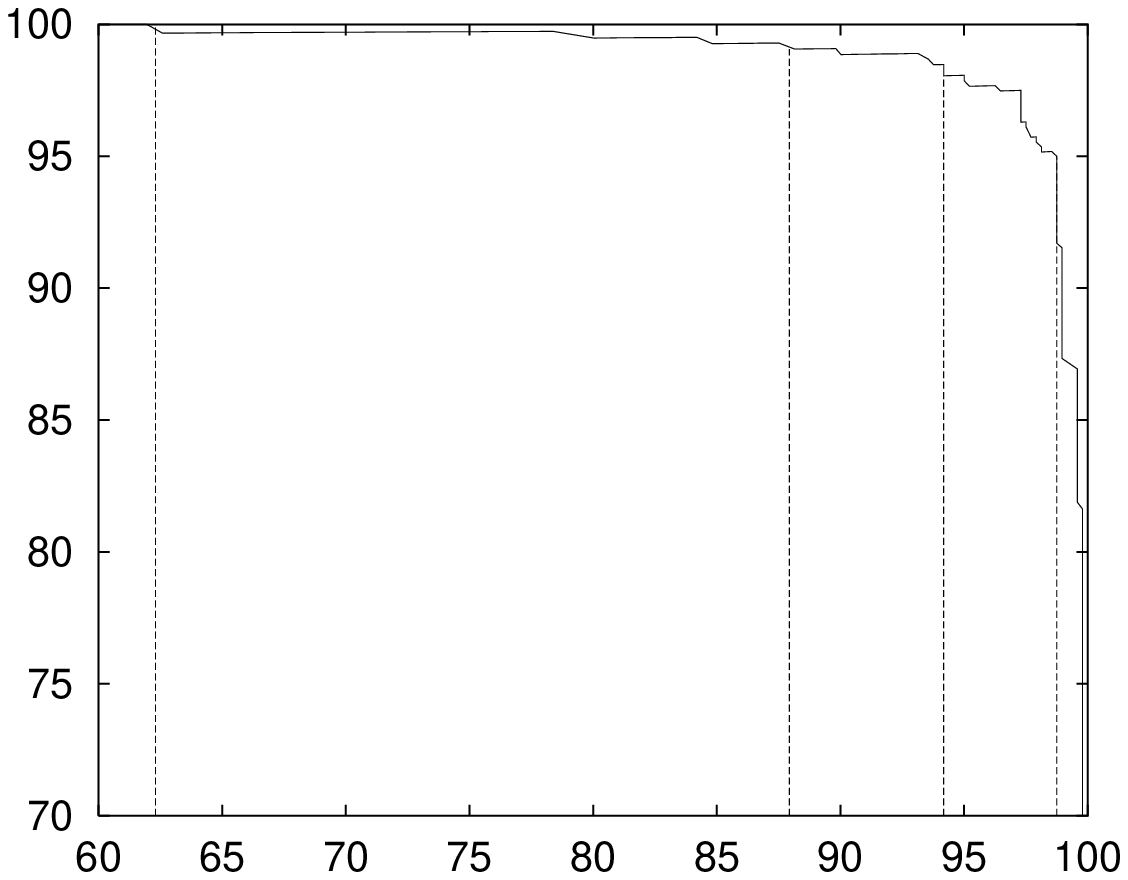, width=4.4cm} &
\hspace*{-0.7cm}\epsfig{file=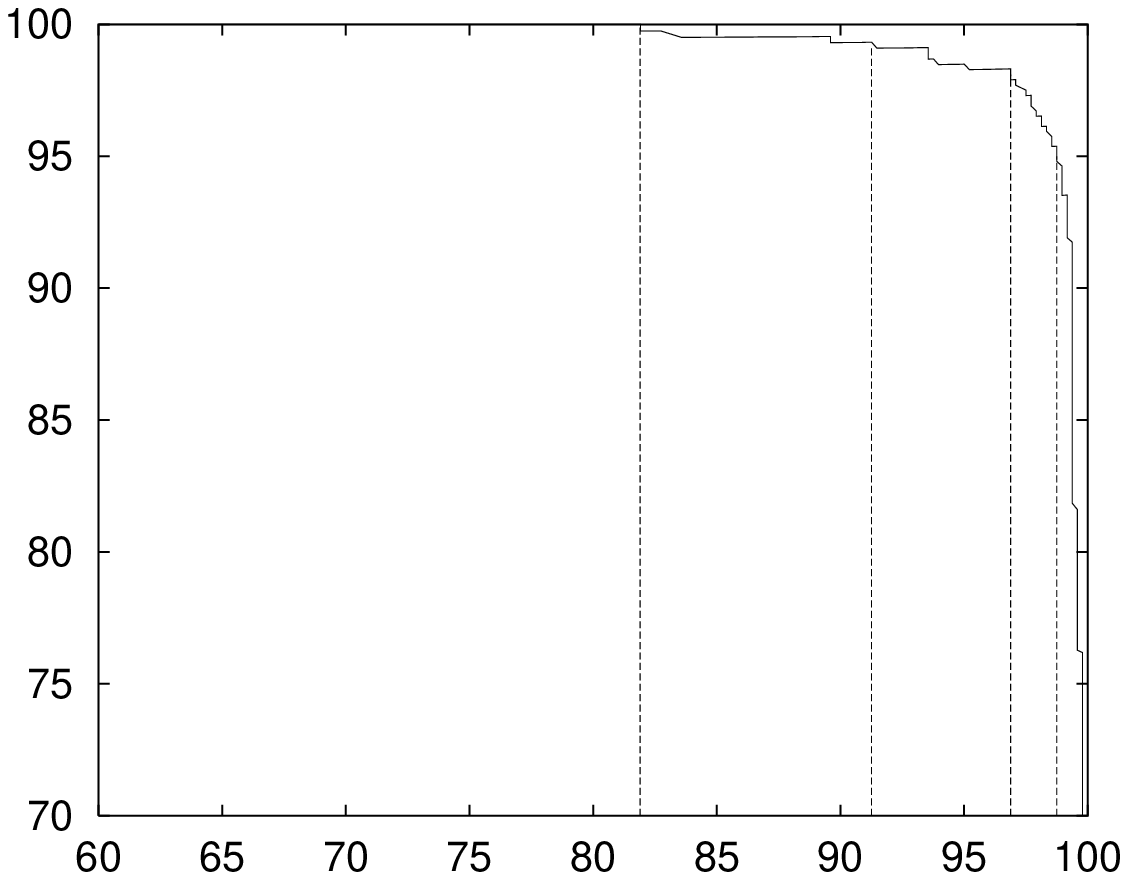, width=4.4cm} \\
\TB{2} & \TB{3} \\
\hspace*{-0.5cm}\epsfig{file=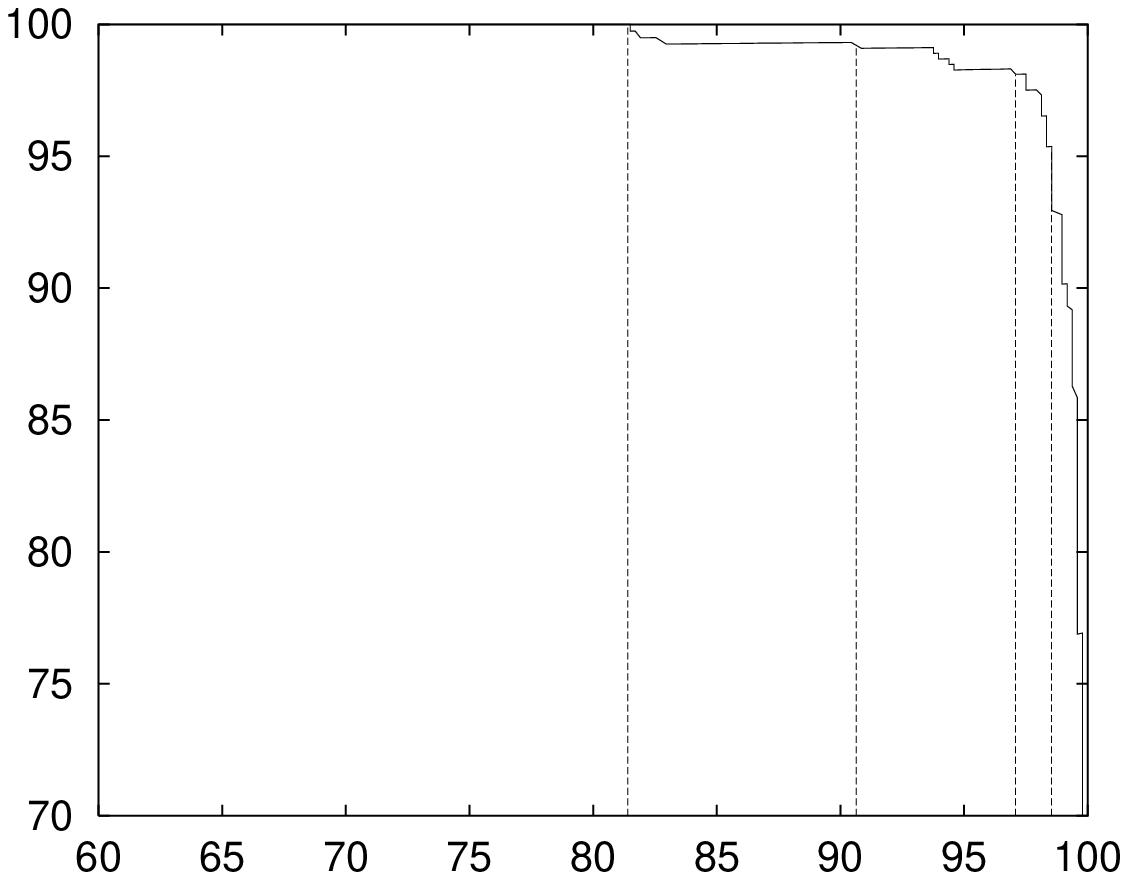, width=4.4cm} &
\hspace*{-0.7cm}\epsfig{file=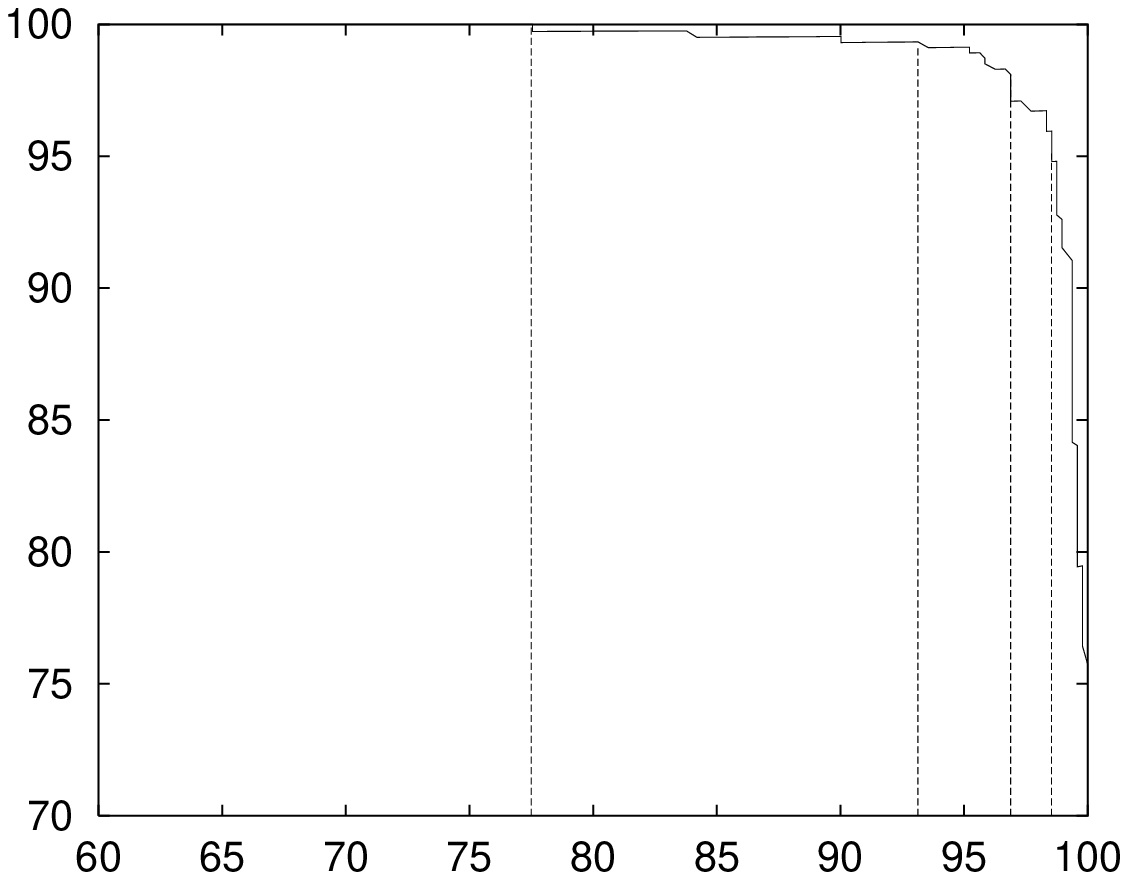, width=4.4cm} \\
\TB{4} & \TB{5} \\
\hspace*{-0.5cm}\epsfig{file=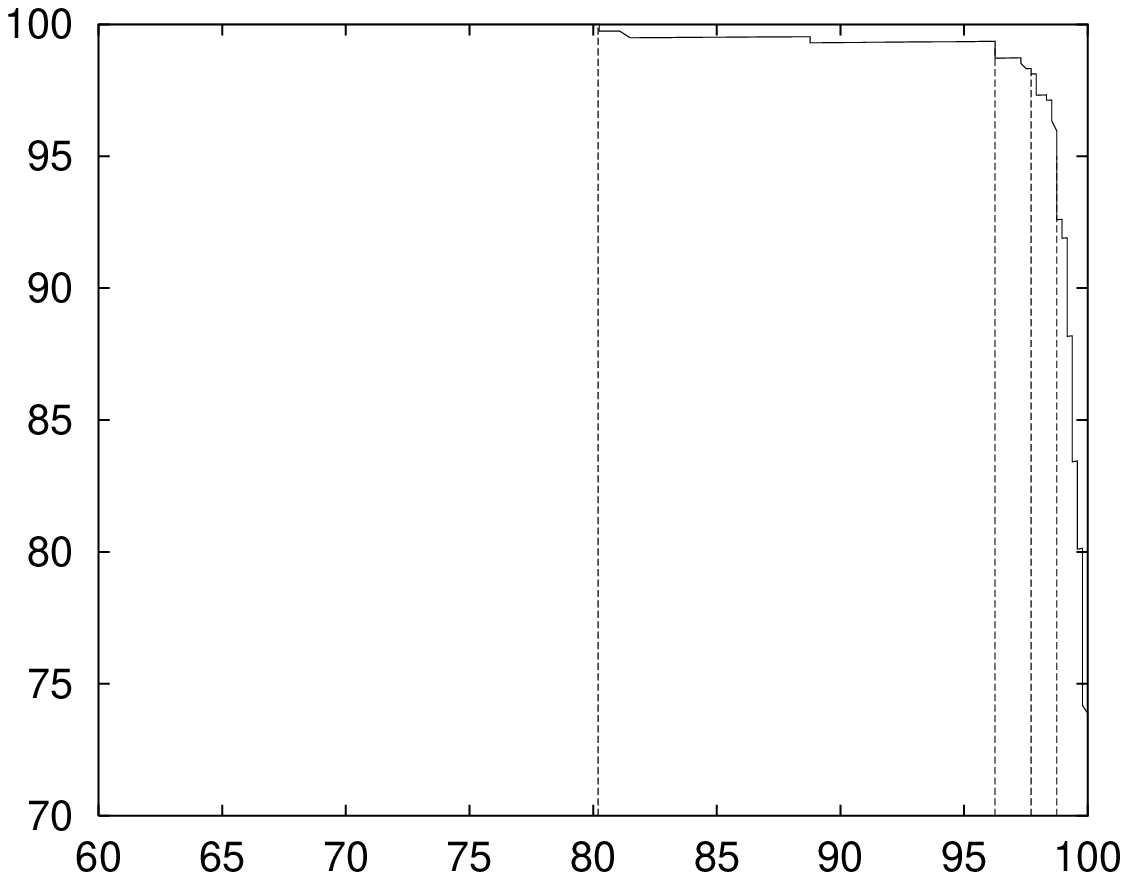, width=4.4cm} &
\hspace*{-0.7cm}\epsfig{file=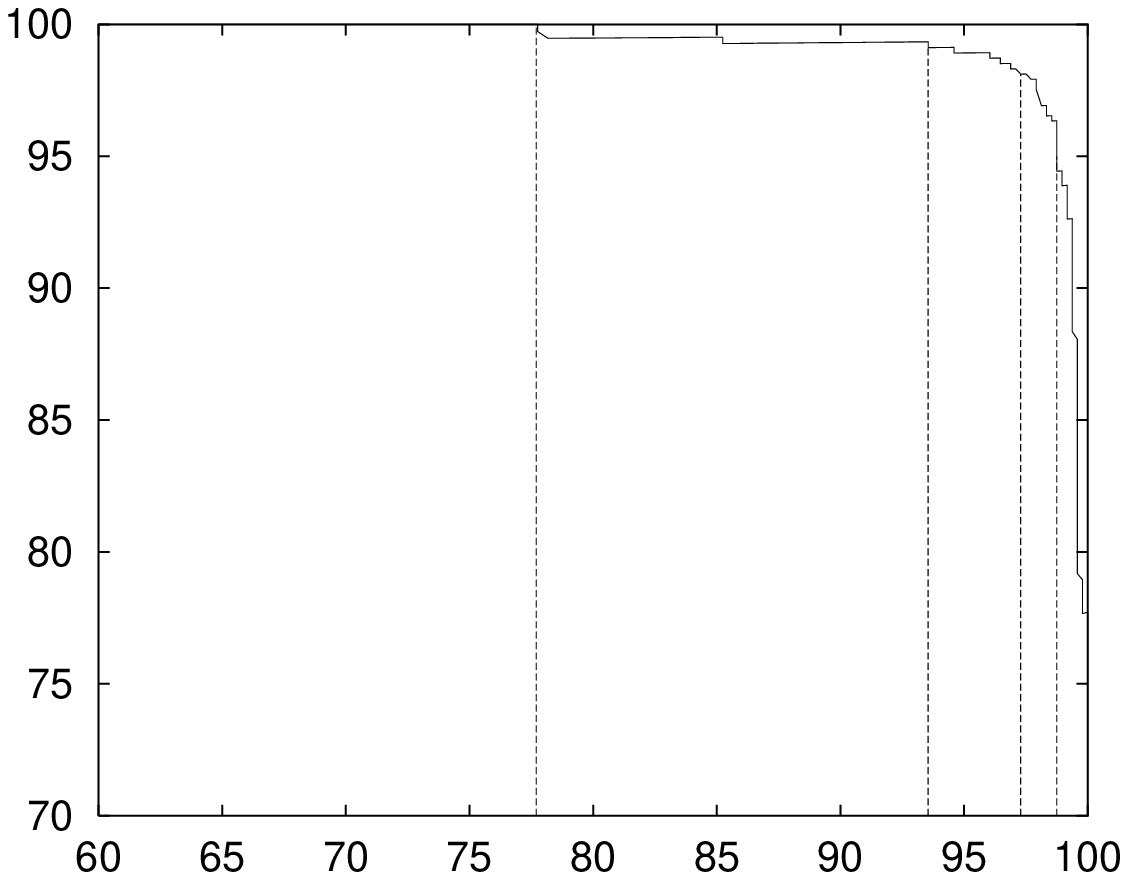, width=4.4cm} \\
\end{tabular}
\caption{Precision-Recall curves and recall values for the fixed
precision rates at 100\%, 99\%, 98\% and 95\%. $x$ axis: recall; $y$
axis: precision.}
\label{fig:securerecall}
\end{figure}

In order to give a clearer picture of the behaviour of classifiers
when moving the decision threshold, we include in Figure
\ref{fig:securerecall} the precision-recall curves of each
classifier. These curves are built giving $\theta$ a wide range of
values, and computing for each value the recall and precision
rates. In these curves, high-precision rates of 100\%, 99\%, 98\% and
95\% have been fixed so as to obtain the recall rate at these points.
Table \ref{tab:securerecall} summarizes these samples.  All the
variants are indistinguishable at level of 95\% of precision. However,
when moving to higher values of precision ($\ge 95\%$) a significant
difference seems to occur between \STUMPS\ and the rest of variants
using deeper weak rules. This fact proves that increasing the
expressiveness of the weak rules can improve the performance when
requiring very high precision filters.  Unfortunately, no clear
conclusions can be drawn about the most appropriate depth. Parenthetically, it can be noted that \TB{4} achieves the
best recall rates in this particular corpus.

\begin{table}[htb]
\centering
\begin{tabular}{|l|c|c|c|c|c|c|c|} \hline
Method & 100\% & 99\% & 98\% & 95\% \\ \hline
\STUMPS & 62.37 & 87.94 &  94.17 & 98.75 \\ \hline
\TB{1} & 81.91 & 91.26 & 96.88 & 98.75 \\ \hline
\TB{2} & 81.49 & 90.64 & 97.08 & 98.54 \\ \hline
\TB{3} & 77.54 & 93.13 & 96.88 & 98.54 \\ \hline
\TB{4} & 80.24 & 96.25 & 97.71 & 98.75 \\ \hline
\TB{5} & 77.75 & 93.55 & 97.29 & 98.75 \\ \hline
\end{tabular}
\caption{Recall rate of filtered spam messages with respect to fixed points of
precision rate}
\label{tab:securerecall}
\end{table}

%\begin{figure*}[htb]
%\centering
%\begin{tabular}{ccc}
%\STUMPS & \TB{1} & \TB{2}  \\
%\epsfig{file=srstumps.525.eps, width=5cm} &
%\epsfig{file=srdepth1.525.eps, width=5cm} &
%\epsfig{file=srdepth2.725.eps, width=5cm} \\
%\TB{3} & \TB{4} & \TB{5} \\
%\epsfig{file=srdepth3.675.eps, width=5cm}    &
%\epsfig{file=srdepth4.450.eps, width=5cm} &
%\epsfig{file=srdepth5.550.eps, width=5cm} \\
%\end{tabular}
%\caption{Precision-Recall curves and recall values for the fixed
%precision rates at 100\%, 99\%, 98\% and 95\%}
%\label{fig:securerecall}
%\end{figure*}

%\subsection{Experiments with other Feature Sets}
%The last experiment of this paper explores the performance of boosting 
%classifiers in the other available versions of the PU1 corpus; these differ in the 
%type of features used to represent the messages. For each version of
%the corpus, namely BARE, LEMM, STOP and LEMM STOP, a \TB{3} classifier 
%has been learned, with up to 2500 rounds of boosting. Since the
%intention of this experiment is to show if more complex
%features can potentially perform better, no validation procedure to
%get a concrete T value is required. Figure \ref{fig:otherfeatures}
%shows the error rate with respect to the used number of weak rules. It 
%can be observed that all classifiers perform very similar (left
%figure), although slight improvements of 1 point can be obtained with
%lemmatized features (right figure). These results agree with the ones
%reported by Androutsopoulos et al. in \cite{androus00b}.

%\begin{figure*}[htb]
%\centering
%\begin{tabular}{cc}
%\epsfig{file=otherfeatures.eps, width=8cm} &
%\epsfig{file=otherfeatures.zoom.eps, width=8cm} \\
%\end{tabular}
%\caption{Error rate of \TB{3} classifiers learned with different
%feature sets}
%\label{fig:otherfeatures}
%\end{figure*}

%----------------------------------------- C o n c l u s i o n s  &
%                                          F u r t h e r    W o r k
\section{Conclusions}
\label{s-conclusions}
The presented experiments show that AdaBoost learning algorithm
clearly outperforms Decision Trees and Naive Bayes methods on the
public benchmark PU1 Corpus. In this data set, the method is resistant
to overfitting and $F_1$ rates above 97\% are
achieved. Procedures for automatically tune the classifier parameters,
such as the number of boosting rounds, are provided.

In scenarios where high-precision classifiers are required, Ada\-Boost
classifiers have been proved to work properly. Experiments have
exploited the expressiveness of the weak rules when increasing their
depth. It can be concluded that deeper weak rules tend to be more
suitable when looking for a very high precision classifier. In this
situation, the achieved results on the PU1 Corpus are fairly
satisfactory.

Two AdaBoost classifiers capabilities have been shown to be useful in
final email filtering applications: a) The confidence of the
predictions suggests a filter which only blocks the more
confident messages, delivering the remaining messages to the final
user. b) The classification threshold can be tuned to obtain a very
high precision classifier.

As a future research line, we would like to study the presented
techniques in a larger corpus. We think that the PU1 corpus is too
small and also too easy: default parameters produce very good results,
and the tuning procedures result only in slight
improvements. Moreover, some experiments not reported here (which study
the effect of the number of rounds, the use of richer feature
spaces, etc.) have shown us that the confidence of classifiers depends
on several parameters. Using a larger corpus, the effectiveness of
the tuning procedures would be more explicit and, hopefully, clear
conclusions about the optimal parameter settings of AdaBoost could be
drawn.

Another line for future research is the introduction of
misclassification costs inside the AdaBoost learning algorithm. Initial 
experiments with the method proposed in \cite{schapire98a} have not
worked properly, although we believe that learning directly
classifiers according to some utility settings will perform better
than tuning a classifier once learned.

%In addition, we would like to experiment with the following two
%learning algorithms: 1) Alternating Decision Trees \cite{freund99b},
%for being a boosting-based technique that generalizes the notion of
%our learned trees, and 2) Support Vector Machines, for being the other
%top-performing method in Text Categorization problems.

%----------------------------------------- A g r a i m e n t s
%\section*{Acknowledgments} %This research has been partially funded
%by the Spanish %Research Department (CICYT's project
%TIC98--0423--C06), by the EU %Comission (NAMIC IST-1999-12392), and by
%the Catalan Research %Department (CIRIT's consolidated research group
%1999SGR-150 %and CIRIT's grant 1999FI 00773).  \vspace*{10mm}

\section*{Acknowledgments}
This research has been partially funded by the EU
(IST-1999-12392) and by the Spanish (TIC2000-0335-C03-02,
TIC2000-1735-C02-02) and Catalan (1997-SGR-00051) Governments. Xavier Carreras holds a grant by the
Department of Universities, Research and Information Society of the
Catalan Government.

\bibliographystyle{ranlp}
\begin{scriptsize}
\bibliography{bibliografia}
\end{scriptsize}

\end{document}